\documentclass[conference]{IEEEtran}
\IEEEoverridecommandlockouts

\usepackage{microtype}
\usepackage{graphicx}

\usepackage{booktabs} 
\usepackage{multirow}
\usepackage{mathtools}
\usepackage{array}
\usepackage{tablefootnote}
\usepackage{comment}
\usepackage{color}
\usepackage{xcolor}
\usepackage{subfig}
\usepackage{amsfonts}
\usepackage{xspace}
\newcommand{\ours}{LL-ViT\xspace}

\begin{document}
\title{LL-ViT: Edge Deployable Vision Transformers with Look Up Table Neurons}

\DeclareRobustCommand{\IEEEauthorrefmark}[1]{\smash{\textsuperscript{\footnotesize #1}}}

\author{
Shashank Nag{\IEEEauthorrefmark{1}}\textsuperscript{*}, Alan T.L. Bacellar{\IEEEauthorrefmark{1}}, Zachary Susskind{\IEEEauthorrefmark{1}}, Anshul Jha{\IEEEauthorrefmark{2}}, \\ Logan Liberty{\IEEEauthorrefmark{1}}, 
Aishwarya Sivakumar{\IEEEauthorrefmark{1}},  Eugene B. John{\IEEEauthorrefmark{2}}, Krishnan Kailas{\IEEEauthorrefmark{3}}, \\ Priscila M.V. Lima{\IEEEauthorrefmark{4}}, 
Neeraja J. Yadwadkar{\IEEEauthorrefmark{1}}, 
Felipe M.G. França{\IEEEauthorrefmark{5}}, Lizy K. John{\IEEEauthorrefmark{1}} \\
{\IEEEauthorrefmark{1}}Department of Electrical and Computer Engineering, The University of Texas at Austin, USA \\
{\IEEEauthorrefmark{2}}The University of Texas at San Antonio, USA   \quad {\IEEEauthorrefmark{3}}Independent Researcher\textsuperscript{$\dagger$} \thanks{\textsuperscript{$\dagger$}Work done while at IBM T.J. Watson Research Center, Yorktown Heights, USA}   \\
{\IEEEauthorrefmark{4}}Federal University of Rio de Janeiro, Brazil \quad
{\IEEEauthorrefmark{5}}Instituto de Telecomunicações, Portugal\\
\textsuperscript{*}Email: shashanknag@utexas.edu 
  
} 

\maketitle






\begin{abstract}

Vision Transformers have been tremendously successful in computer vision tasks. However, their large computational, memory, and energy demands are 
a challenge
for edge inference on FPGAs -- a field that has seen a recent surge in demand. We recognize the benefits of recent works on logic and Look Up Table (LUT) based networks, such as LogicNets, NeuraLUT, DWN, among others, in offering models that simultaneously reduce both the memory and compute footprints. However, these models natively do not perform well on common vision tasks, such as CIFAR-10/100. In this work, we propose \ours, a novel edge optimized vision transformer design that integrates layers of LUT neurons within the transformer architecture. 
Based on our characterization that reveals that a majority of model weights and computations are from the channel mixer (MLP layer), we design an alternate LUT-based channel mixer, and simultaneously develop an FPGA-based accelerator for \ours. 
Contrary to some attempts to replace each multiplication with a table lookup, our architecture utilizes a neural learning approach which natively learns the LUT functions.
This approach allows for reduced model sizes, and a computational and energy-efficient inference solution for vision transformer models. Evaluating on edge-suitable workloads, we achieve accuracies of 95.5\% on CIFAR-10, 78.8\% on CIFAR-100, and 60.9\% on Tiny-ImageNet datasets, comparable to the baseline transformer. 
\ours eliminates over 60\% of the model weights and 50\% of the multiplications in the model, and achieves 1.9$\times$ energy efficiency and 1.3$\times$ lower latency over an integer quantized ViT accelerator, while also offering superior throughput against prior works at a 10.9W power budget. 

\end{abstract}

\section{Introduction}\label{sec:intro}


Vision Transformers (ViTs) have recently garnered significant attention due to their versatility and strong performance across a wide range of vision tasks, including image classification, object detection, and semantic segmentation~\cite{vit, deit, vit-segmentation, imagenet-leaderboard}. With the rise of generative models such as DALL-E~\cite{dall-e}, the importance of ViTs has only grown, positioning them as a key component in modern computer vision systems.

There is a growing demand for deploying computer vision models on edge devices, driven by applications in robotics, autonomous systems, and other latency-sensitive edge domains. While convolutional neural networks (CNNs) have traditionally dominated this space, a recent market trend highlights increasing interest in deploying ViTs at the edge~\cite{sld-report}. Many ViTs exhibit regular inter-layer structure, which offers opportunities for efficient dataflow design and hardware acceleration. Furthermore, as ViT architectures continue to evolve rapidly, FPGAs 
present a promising platform 
for their deployment owing to their flexibility and customizability.

Despite their potential, deploying ViTs on FPGAs for edge inference remains challenging. These models are large and computationally intensive, with memory and energy requirements that far exceed the capabilities of commercial edge platforms~\cite{vita}. Even compact variants such as DeiT-T~\cite{deit} require $\sim$~20MB of weight storage, equivalent to $\sim$~9100 BRAM18s, surpassing the on-chip memory available on many FPGAs.

To address these limitations, prior works have explored various optimization strategies, including on-demand loading of weights from off-chip memory and quantization-based compression~\cite{vita,mevit,autovitacc}. However, both approaches have trade-offs: frequent off-chip memory access significantly increases energy consumption, while aggressive quantization may degrade model accuracy. More recently, hybrid solutions using quantized models combined with pipelined execution on platforms like the AMD/Xilinx Versal AI Engine have demonstrated high throughput \cite{ssr, hgpipe}, but often at the cost of high power consumption (of the order of $40$W), which is unsuitable for battery-operated, power-constrained edge scenarios.

These challenges motivate the exploration of alternative paradigms that simultaneously reduce both model size and compute requirements. Recent work in look-up-table (LUT) and logic-based neural networks, including PolyLUT~\cite{polylut}, NeuraLUT~\cite{neurallut},  Differentiable Weightless Neural Networks (DWNs)~\cite{dwn}, LogicNets~\cite{logicnets}, amigoLUT~\cite{amigolut}, and treeLUT~\cite{treelut}, has shown promise in leveraging compact representations with efficient lookup-based inference. DWNs, in particular, introduce differentiable mechanisms for learning LUTs using approximate gradients, enabling compact models with massive potential for hardware efficiency. However, such models have thus far been limited to small datasets and simpler tasks, and their performance on standard computer vision benchmarks like CIFAR-10 remains relatively low~\cite{dwn}.

In this paper, we investigate how LUT-based neural layers can be integrated into the architecture of ViTs to make them more suitable for edge deployment, specifically targeting edge-suitable workloads such as CIFAR-10, CIFAR-100~\cite{cifar10} and TinyImageNet~\cite{tinyimagenet}. Our goal is to co-design models and hardware architectures that enable real-time inference while reducing both memory footprint and energy consumption.

\begin{figure}[htbp!]
  \centering
  {\includegraphics[width=0.4\textwidth, keepaspectratio]{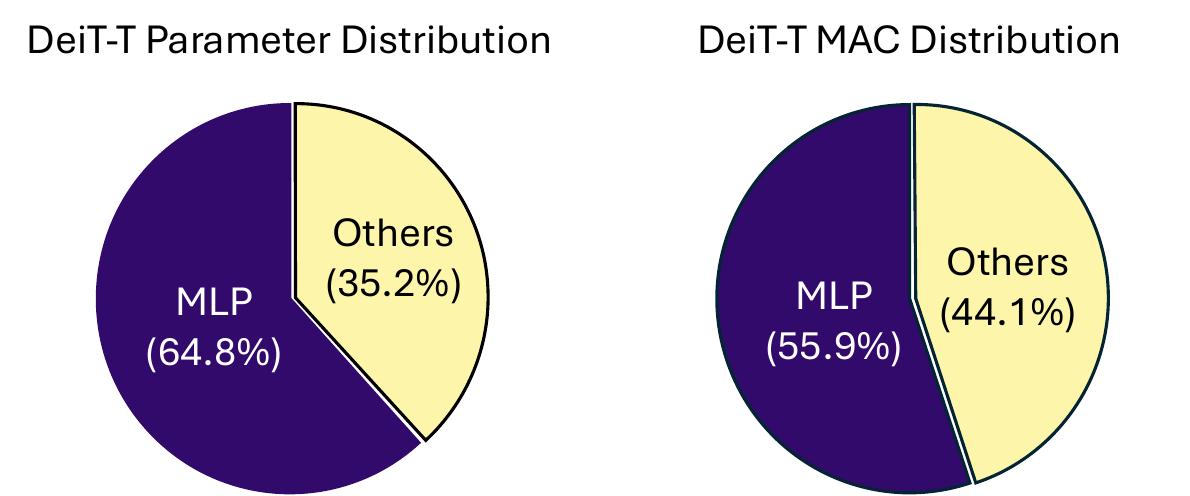}}
  \caption{The MLP layers contribute over 60\% of the overall model weights and 55\% of the overall multiply-accumulate (MAC) operations in the DeiT-T vision transformer \cite{deit}.}
  \label{fig:intro_pie}
\end{figure}

In vision transformers, multi-head self-attention (MHA) layers are generally used for token mixing, and Multilayer Perceptron (MLP) layers are used for channel mixing~\cite{channelmixer}. 
We observe that in typical ViTs, a major fraction of compute and model weights is attributed to the MLP layers, 
as shown in Fig.~\ref{fig:intro_pie}. 
While various model optimization techniques with transformers have been explored in the past, the core MLP block has continued to dominate the overall model, and it has been shown that these MLPs are infact the key behind the knowledge learnt in these models~\cite{mlp-knowledge2}. 

This observation motivates us to explore a LUT-based neural network design to implement channel mixing in vision transformers, in lieu of the MLP block. 
Look-up-Table (LUT) or RAM node based neurons, which involve no multiplications, and only look-up operations, offer a much more energy-efficient inference solution than traditional multiply-add neurons~\cite{bthowen, logicnets,dwn}. At implementation, the LUT-neurons can be effectively packed into the LUT slices on FPGAs. These neuron implementations are self-contained, and can operate without additional compute or memory resources required for storing weights or to generate outputs -- saving BRAMs, DSPs and LUT slices for other compute, thus yielding high energy savings. 

We also note that learning LUT-neurons from scratch would lead to smaller and efficient structures than performing a post-training mapping of conventional neurons, or individual multiplications to LUTs~\cite{mwscas, dwn}. {{While multiplication free networks that convert multiplications into table lookups have been proposed, each multiplication would end up being converted to one table lookup~\cite{lut-nn}. 
\begin{figure}[htbp!]
  \centering
  {\includegraphics[width=0.4\textwidth, keepaspectratio]{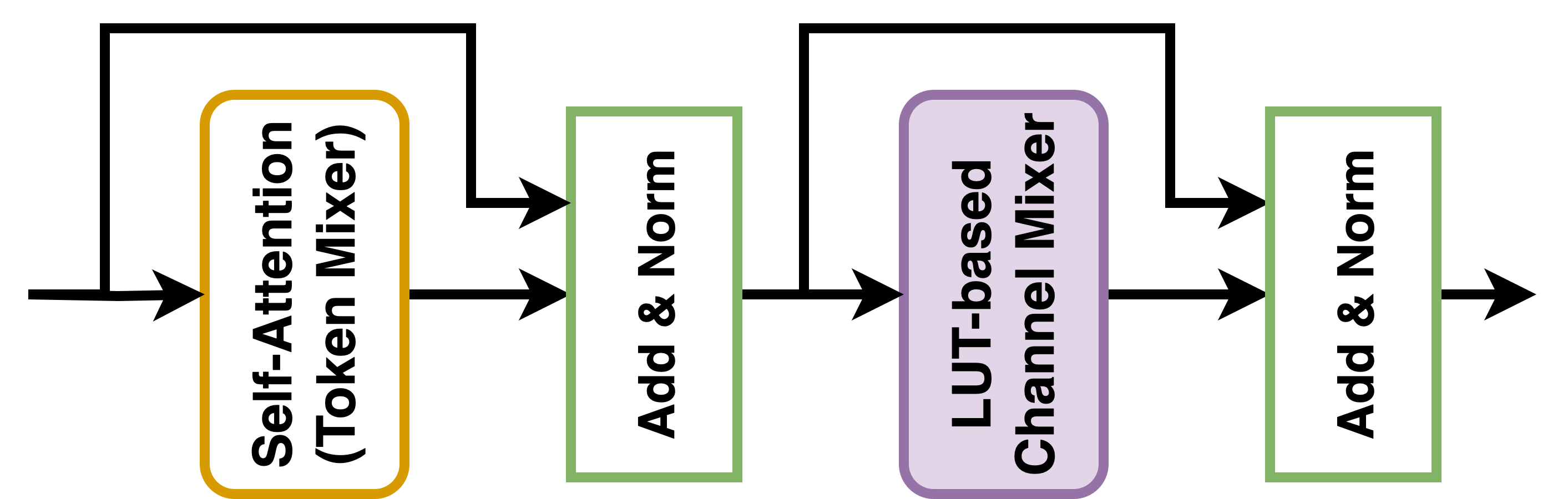}}
  \caption{Proposed Learned-LUT based Vision Transformer (LL-ViT) - overview of a single encoder layer in the model}
  \label{fig:intro}
\end{figure}

In this work, we  
propose a novel learnable LUT-based channel-mixing block and integrate into a transformer encoder layer(Fig. \ref{fig:intro}). With this, we propose \textbf{\ours} : \underline{L}earned-\underline{L}UT based \underline{Vi}sion \underline{T}ransformers -- an algorithm hardware co-design technique for FPGA-based edge-deployable vision transformers. We offer a class of models optimized for FPGA acceleration, that incorporate multiplication-free channel mixers in vision transformers (Fig. \ref{fig:intro}), 
offering an edge-efficient inference solution that delivers the model performance advantages of conventional vision transformers, while offering hardware performance benefits of LUT neurons. 
Specifically, we make the following contributions in this paper:
\begin{itemize}
     \item We propose a novel learnable LUT-based channel-mixer block, {and demonstrate its seamless integration into a transformer model}.
    \item We integrate the proposed channel-mixer with a self-attention based token mixer, offering an edge-optimized vision transformer model well-positioned for FPGAs, with reduced computational demand and memory consumption. 
    \item We co-design a low-power accelerator for \ours that avoids off-chip weight movement, and evaluate its performance on FPGA against  prior works. 
    \item While logic and LUT-based models have traditionally been applied to small-scale datasets, for the first time, we extend it to vision transformers and demonstrate their performance on relatively complex computer vision tasks. \ours achieves comparable accuracies of 95.5\% on CIFAR-10, 78.8\% on CIFAR-100, and 60.9\% on Tiny-ImageNet, at a smaller model size and computational cost compared to baseline ViTs. 
\end{itemize}
We note that low energy, low latency, and reduced memory requirements with a LUT based implementation of the channel mixer make LL-ViTs excellent candidates for edge inference deployments. LL-ViT offers $1.9\times$ improved energy efficiency and $1.3\times$ lower latency against a quantized baseline FPGA accelerator. With over $60\%$ reduction of model weights, LL-ViT fits completely on a Virtex platform offering a high throughput of $1083$ FPS, at a reasonable power consumption of $10.9$W. 

\section{Background and Related Work} \label{sec:background}
\subsection{Vision Transformers and their FPGA acceleration}

Vision Transformers (ViTs) are models for computer vision applications with a stack of encoders (Fig. \ref{fig:vit}), that were inspired by the success of transformer models for language tasks~\cite{vit, deit, vitg}. Over the years, there has been active research in making ViTs more efficient, both in terms of model performance, as well as 
inference efficiency~\cite{efficient-vit}. However, the core backbone of the encoders in these models is largely similar -- a multi-head self-attention block that captures flow of information across tokens (token mixer), and a Multi-Layer Perceptron (MLP) block to cater to the interaction across different feature channels within each token (channel mixer). Within these blocks, the operations primarily involve matrix multiplications, and non-linear operations including GELU, LayerNorm, among others. 

\begin{figure}[htbp!]
  \centering
  \includegraphics[width=0.45\textwidth, keepaspectratio]{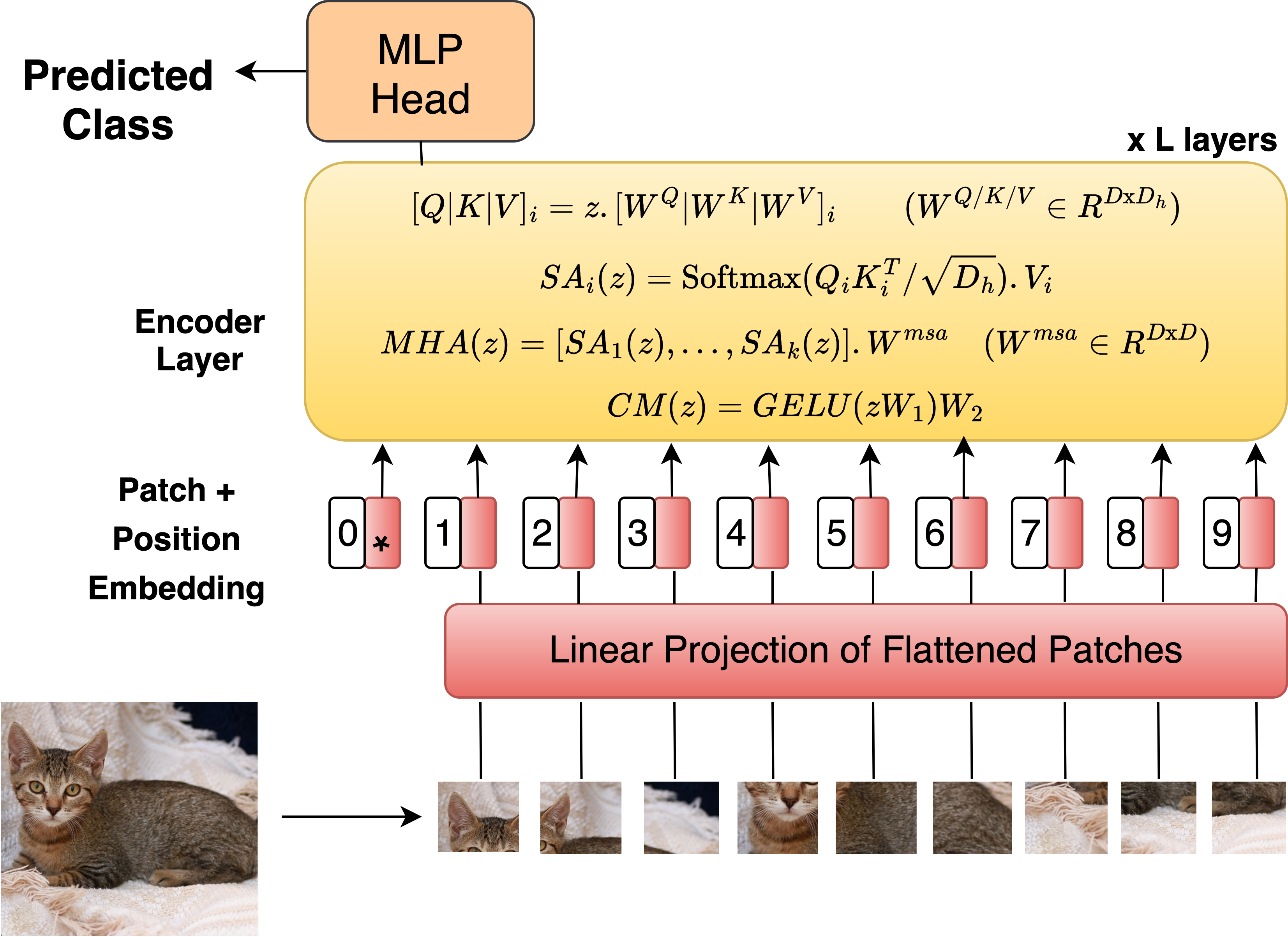}
  \caption{A typical vision transformer model consisting of a stack of encoder blocks. Figure adapted from \cite{vit}. 
  }
  \label{fig:vit}
  \vspace{-10pt}
\end{figure}

In order to address the issues with increasing model sizes and the computational complexity of ViTs, and the consequent deployment implications, various model optimization techniques have been studied in the past. 
Particularly, quantization of these models has been investigated, which helps reduce the memory requirement and compute precision. 
I-ViT \cite{ivit} and I-BERT \cite{ibert} quantize weights and activations to 8-bits, and BinaryViT \cite{binaryvit} has explored binary-precision quantized designs that are suitable for efficient inference implementations. However, the computational complexity of these models remains unchanged. Compact Vision/ Convolution Transformers \cite{cct} proposes a set of compact ViTs with fewer encoder layers, that are well-suited for the small-data and resource-constrained environment regime. SwiftFormer proposes an additive attention mechanism geared for mobile applications, as an alternative to the self-attention mechanism~\cite{swiftformer}. 

In recent years, numerous works have explored energy-efficient transformer architectures and their hardware accelerator designs targeting FPGAs. ViTA~\cite{vita} introduced an edge-optimized accelerator, and ME-ViT~\cite{mevit} developed an accelerator that minimizes memory traffic; but both fetch in weights from off-chip memory atleast once every inference request. ViA~\cite{via} proposes an FPGA architecture with an effective partitioning strategy, but involves write backs to memory between the MLP and MHA processing blocks. HeaT-ViT~\cite{heatvit} offers an algorithm-hardware co-design solution using token selectors and pruning strategies to reduce the computation. More recent works include SSR~\cite{ssr} and HG-PIPE~\cite{hgpipe} primarily targeting Versal platforms. While SSR~\cite{ssr} uses a combination of temporal and spatial acceleration to offer higher throughput or latency, HG-PIPE adopts an aggressive quantization (3-bit / 4-bit) and LUT-based compute techniques to ensure that the model weights can reside on-chip. However, both of these works suffer from high power consumption, making them sub-optimal for the edge. 

\begin{table*}[ht!]
\centering
\begin{tabular}{ ccccc }
 \toprule
 Layer & \# Parameters & \# MAC ops & DeiT-T : \# Parameters & DeiT-T : \# MAC ops\\
 \midrule
 Q, K, V Projection & $3\times n_{layers} \times D \times D$ &  $3  \times n_{layers} \times N \times D \times D$  & 1,327,104 & 261,439,488 \\
 $Q.K^T$ & - & $ n_{layers} \times N \times D \times N$ & - & 89,415,936 \\
 SoftMax.V & - & $ n_{layers} \times N \times N \times D$ & - & 89,415,936 \\
 Multi-head concat & $ n_{layers} \times D \times D$ & $n_{layers} \times N \times D \times D$ & 442,368 & 87,146,496\\
 \midrule
 Feed-forward 1 (MLP) & $4 \times n_{layers} \times D \times D$  &  $4\times n_{layers} \times N \times D \times D$ & 1,769,472 & 348,585,984\\
 Feed-forward 2 (MLP) & $4\times n_{layers} \times D \times D$  &  $4\times n_{layers} \times N \times D \times D$ & 1,769,472 & 348,585,984 \\
 \bottomrule
\end{tabular}
\vspace{5pt}
\caption {Breakdown of computations and model weights in a typical Vision Transformer. $n_{layers}$ represents the number of layers, $N$ represents the number of tokens, and $D$ is the latent dimension of the model. In DeiT-T, $n_{layers}=12$, $D=192$, for an input image of size $224$ $\times$ $224$, with a patch size of $16$ $\times$ $16$, and one token for the classification head, $N=197$. 
}
\vspace{-5pt}
\label{table:workload}
\end{table*}





\subsection{Multiplication-free Neural Networks}\label{sec:background-amm}

Of late, there has been a surge of interest in approximate matrix multiplication (AMM) to enhance energy efficiency of model inference. 
A transformer architecture was proposed that approximates floating-point multiplications with piece-wise affine functions, achieving comparable performance while reducing computational demands primarily for energy savings~\cite{affine}. 
LUT-GEMM~\cite{lut-gemm-iclr} introduces an AMM method using Look-up-Tables (LUTs) to reduce both energy consumption and latency by modifying GPU kernels. LUT-NN~\cite{lut-nn} utilizes a centroid-based multiplication approximation technique to replace multiplications with lookup operations.
However, these methods primarily rely on LUTs as an inference-time optimization, substituting multiplication operations with lookup operations to achieve post-training speedup. We now turn our attention to 
an unconventional class of neural networks 
that employs LUTs as the core computational units or ``neurons" of the model, learning them directly during training. This integration of LUTs at the neural architecture level goes beyond inference optimizations, aiming to fully utilize LUT's learning capacity in the training process for improved efficiency.

\noindent \textbf{Learned LUT/ LUT-neuron based Neural Networks: }
In recent times, LUT-neuron based neural networks have gained significant traction. These neurons (Fig.~\ref{fig:neuron}) are efficient to implement, and eliminate the need for power and resource hungry multiply-accumulate (MAC) operations in conventional neurons. 
\begin{figure}[htb!]
\centerline{\includegraphics[width = 7cm]{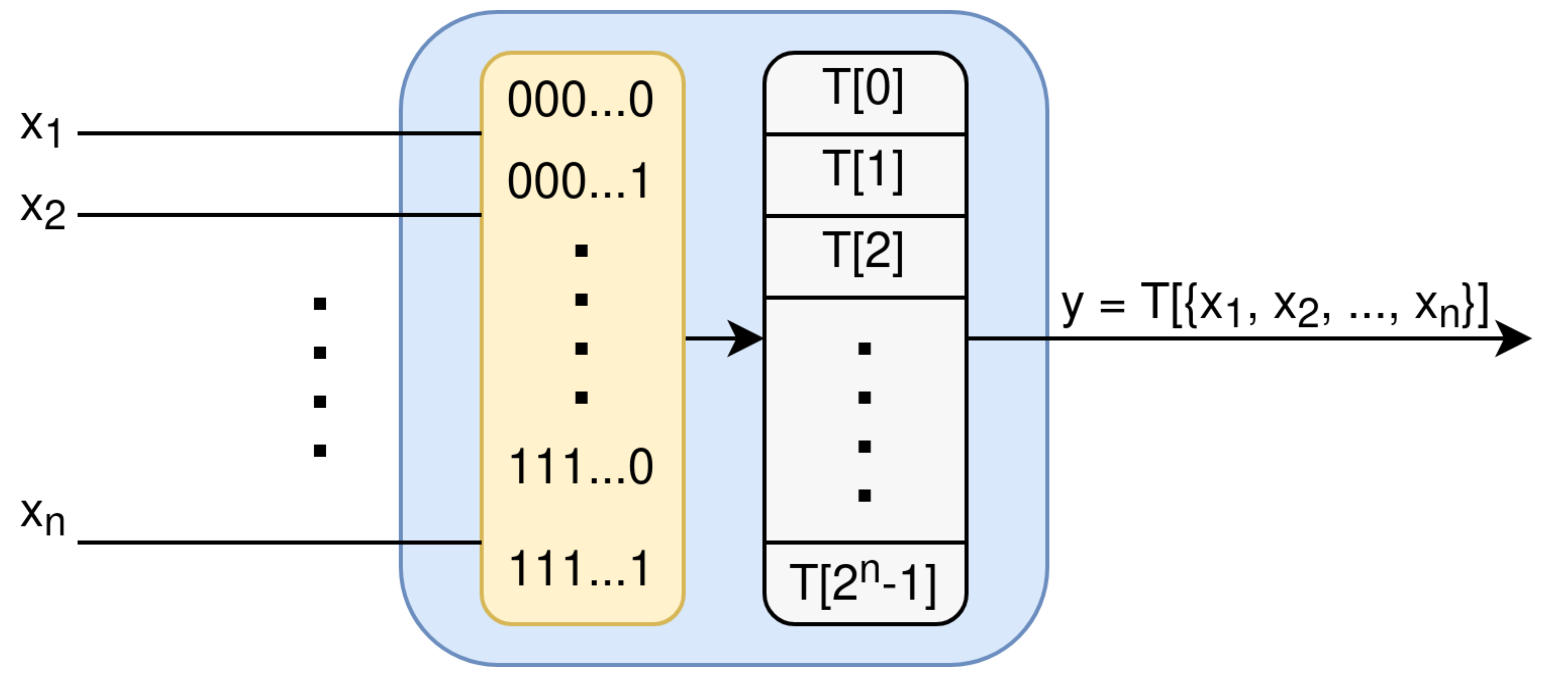}}
\caption{
LUT or RAM-node Neuron : The input sequence is concatenated and used to "look up" the output in the LUT -- with no MAC operations involved.}
\label{fig:neuron}
\end{figure}
LogicNets~\cite{logicnets}, an early work in this realm, trained models using quantized linear neurons that are sparsely connected across layers -- each of these would then be translated to a LUT at implementation. PolyLUT~\cite{polylut} extends this idea, by encapsulating neurons which can learn polynomial functions as LUTs. NeuraLUT~\cite{neurallut} further extends this to support new dense subnetwork architectures encapsulated within each logical-LUT. CompressedLUT~\cite{compressedlut} and ReducedLUT~\cite{reducedlut} offer solutions to efficiently compress these logical-LUTs for their hardware implementation using physical-LUTs. TreeLUT~\cite{treelut} trains Gradient Boosted Decision Trees (GBDTs), and converts them to efficient FPGA-based implementations for efficient inference. NullaNet~\cite{nullanet} is another work which trains networks with binary activations and floating point weights, enabling layers to be interpreted as boolean functions that can then be reduced for implementation.  
DiffLogicNets~\cite{petersen2022} designs differentiable logic-based neural networks, which can be implemented as logic gates (which are effectively LUT2s).
Weightless Neural Network (WNN) is another such class of neural networks inspired by the dendritic trees of biological neurons \cite{wisard, wisard_fpga}. 
These networks train LUTs natively using approximate gradient and estimation-based techniques, based on the idea that an $n$-input LUT is a highly expressive structure which can represent any one of $2^{2^n}$ possible non-linear functions. Consequently, training a LUT-neuron directly could be more efficient than representing a neuron using a LUT, as a LUT-neuron has a higher learning capacity, with a VC dimension of $2^n$ \cite{vc-dimension}, compared to a conventional DNN neuron, which has a VC dimension of $n + 1$. 
Recent works such as BTHOWeN~\cite{bthowen} and ULEEN~\cite{uleen} have demonstrated models with shallower layers and fewer neurons compared to traditional DNNs, at comparable accuracies. 
Differentiable Weightless Neural Networks (DWN)~\cite{dwn}, a recent work in this direction, defined Extended Finite Difference based gradients for LUT-entries and inputs, allowing models with multiple layers of LUT neurons to be differentiable and trained using backpropagation. 
%
All these LUT-neuron based models significantly excel in terms of model size, latency and energy efficiency, as demonstrated on small-scale datasets including JSC~\cite{jsc}, MNIST~\cite{mnist} and NID~\cite{unswnb15}. 
However, these models perform poorly on image classification tasks such as the CIFAR-10 dataset~\cite{cifar10}, as these do not implement an architecture that can learn positional independence of features. 
While these prior works were largely tiny discriminator-based models that directly predicted the maximum likelihood class when presented with an input image, in this work we seek to design LUT-based channel mixers, as intermediate layers within a complex model architecture. 

\section{\ours : Learned-LUT Vision Transformers} \label{sec:design}
\subsection{Workload Analysis \& Motivation}\label{sec:insights}

We study a range of vision transformer models to characterize the computations and memory usage in their constituent layers with an aim to identify bottlenecks in model inference efficiency, and summarize the findings in Table \ref{table:workload}. 
As illustrated here, and as alluded in Fig. \ref{fig:intro_pie}, particularly of interest is the fact that across vision transformer models, channel mixers (MLPs) contribute over 60\% of the total model weights. At the same time, these blocks also contribute to a significant fraction of the overall compute -- ranging between 50-70\% of the total MAC (multiply-accumulate) operations. These findings are also corroborated in ViTA~\cite{vita}. 
This analysis motivates us to look at alternate layer architectures to implement channel mixing in vision transformers, as an efficient channel mixer would translate to significant overall efficiency improvements.
As compared to MLP-only models, prior works on LUT-neuron based models have demonstrated iso-accuracy performance with significant hardware performance improvements~\cite{uleen,dwn}. This suggests that a newly designed LUT-based channel-mixer would be able to learn the patterns originally represented by the MLP block, thus motivating our design. 
For this work, we choose to adopt a WNN-based approach for the LUT-based channel-mixer design. This is owing to the LUT differentiation techniques proposed in DWN~\cite{dwn}, which would enable us to construct a multi-layer LUT-based network with arbitrary connections. The superior performance of DWNs further motivates our choice. 

\subsection{LUT-based Channel Mixer}\label{sec:channelmixer}

\begin{figure*}[htb!]
\centering
 {\includegraphics[width=\textwidth, keepaspectratio]{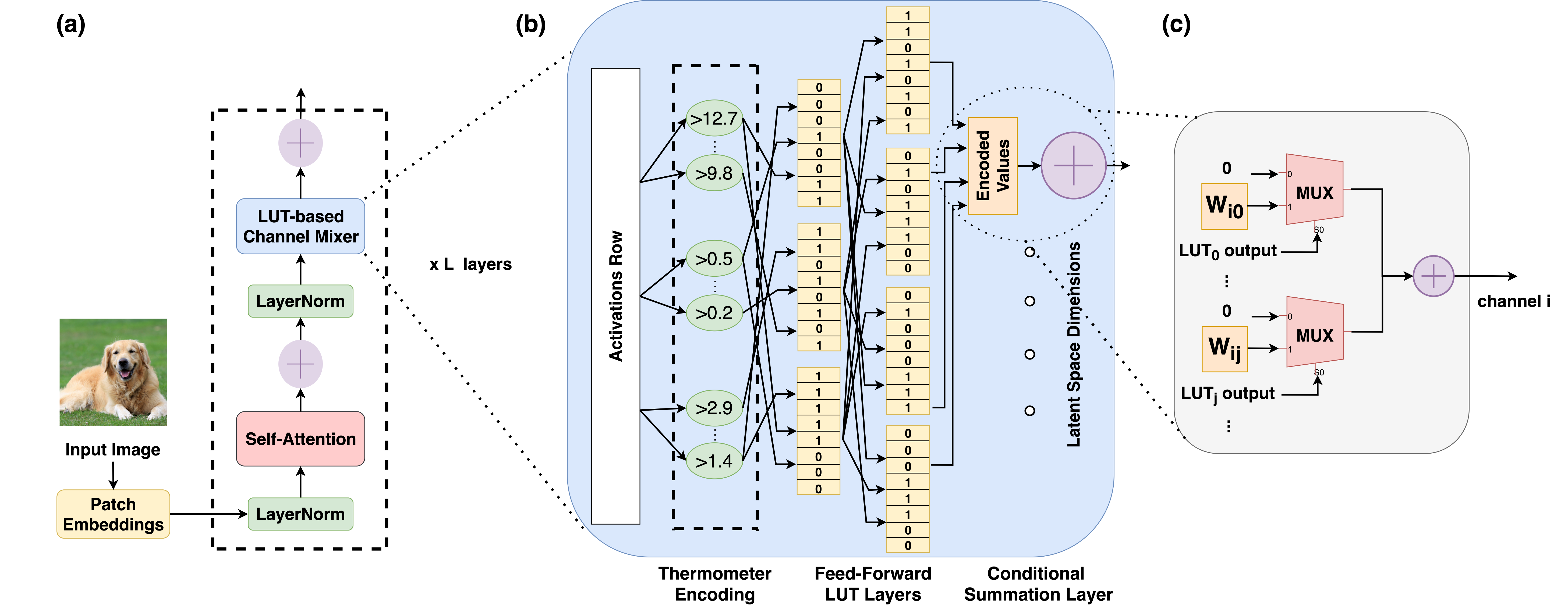}}
 \caption{Proposed Learned LUT-based Vision Transformer Design -- (a) Overall Design, (b) a LUT-based Channel Mixer within the encoder block, 
 (c) the conditional summation layer implementation for a particular channel. This is repeated in each encoder.} 

 \label{fig:weightless_block}
\end{figure*}

\noindent\textbf{Existing Implementation:} The channel mixer transforms the feature channels while treating each token independently. In vision transformers, this is typically implemented as a two-layer MLP with an intermediate GELU non-linearity:
\begin{equation}
    \mathbf{CM}(\mathbf{z}) = \text{GELU}(\mathbf{z} \mathbf{W}_1) \mathbf{W}_2
\end{equation}
where $ \mathbf{z} \in \mathbb{R}^{N \times D} $ is the input token feature (activation) matrix, with $ N $ tokens and $ D $ feature dimensions. The first linear transformation weight matrix is $ \mathbf{W}_1 \in \mathbb{R}^{D \times D_h} $, which expands the feature dimension to $ D_h $. The GELU activation function is applied element-wise after this transformation. Finally, $ \mathbf{W}_2 \in \mathbb{R}^{D_h \times D} $ projects the hidden representation back to the original feature dimension.

\noindent\textbf{Design Challenges:} As mentioned previously, prior work on LUT-based WNNs have proposed models that directly compute the scores for discriminators corresponding to each class. On the contrary, within the vision transformer architecture, channel mixer is an intermediate block, and hence the proposed LUT-based channel mixer would have to present real-valued outputs for all the feature dimensions -- while also ensuring that the output layer is fully-differentiable. 
To address this, we propose a conditional summation layer following the final layer of LUT neurons, that adds higher precision encoded values based on the output requirements from the block. 


\noindent\textbf{Conditional Summation Layer:} We propose a conditional summation layer that operates as follows: if the output from the LUT neuron in the last layer is a $1$, an encoded value is added to the corresponding output, and if it is $0$, it is simply skipped: 
\begin{equation}
    y_i = \sum_{j} 
\begin{cases} 
W_{ij} , & \text{if } x_{j} = 1 \\
0, & \text{otherwise}
\end{cases}
\end{equation}
where \( y_i \) is the output of the \( i \)-th channel, \( x_{j} \) is the output of the \( j \)-th LUT in the last layer, and \( W_{ij} \) represents the corresponding encoded value associated with the \( j \)-th LUT for the \( i \)-th channel. The summation ensures that only the value \( W_{ij} \) corresponding to active LUT outputs (\( x_{j} = 1 \)) contribute to \( y_i \).
Importantly, this layer incurs 
no multiplication during inference, as it merely adds an encoded value based on which LUTs participate in the summation -- while still maintaining the full-precision required by the model. 
While training, we allow the encoded values (\( W_{ij} \)) to be learned with full-precision (fp32) to ensure it is completely differentiable and has smooth gradients. Post-training quantization is performed on these encoded values to quantize them down to a lower precision as required by the rest of the vision transformer. 

\noindent\textbf{Design:} Fig. \ref{fig:weightless_block} (b) illustrates the proposed LUT-based channel mixer block. 
As the channel mixer only captures interactions along the channel, and treats each token independently, 
we flatten input activations row-wise and pass them through these layers one row at a time. We add a thermometer encoding layer~\cite{thermometer, dwn}, which converts the input activations into a sequence of bit representations. This is followed by one or more configurable LUT neuron layers (feed-forward), ending with the conditional summation layer. The output layer contains a number of summation units matching the number of channels in the transformer (latent dimension, $D$), preserving the network’s latent space dimensionality. As discussed earlier, we employ the Extended Finite Difference based technique from DWN~\cite{dwn} for the LUT neurons to be differentiable. This coupled with the differentiable conditional summation layer ensures that the proposed multiplication-free channel mixer is fully learnable, differentiable and compatible with the ViT model architecture. 
\subsection{Integrated Learned LUT Vision Transformer}
As the token mixer block is not dominated by weights, and predominantly involves finding relations amongst the tokens, a LUT-based neural layer for it would not yield significant returns. As such, we stick to existing multi-head self-attention based token mixer blocks, and combine them with our proposed channel mixer block to deliver \ours. 
In doing so, we leverage the LUT neurons' hardware and energy savings advantages combined with the ability to learn positional invariance through the self-attention layers of the transformers. 
We note that we integrate the proposed channel mixer into each encoder in the vision transformer. With the gradients well defined for the LUTs and the conditional summation layer, we ensure that the overall model is fully-differentiable, and the entire \ours is trained end-to-end by backpropagating the losses from the output -- similar to how a regular vision transformer is trained.   
Fig.~\ref{fig:weightless_block}~(a) shows an overview of the proposed design. 

\begin{figure*}[htbp!]
  \centering
  {\includegraphics[width=0.95\textwidth, keepaspectratio]{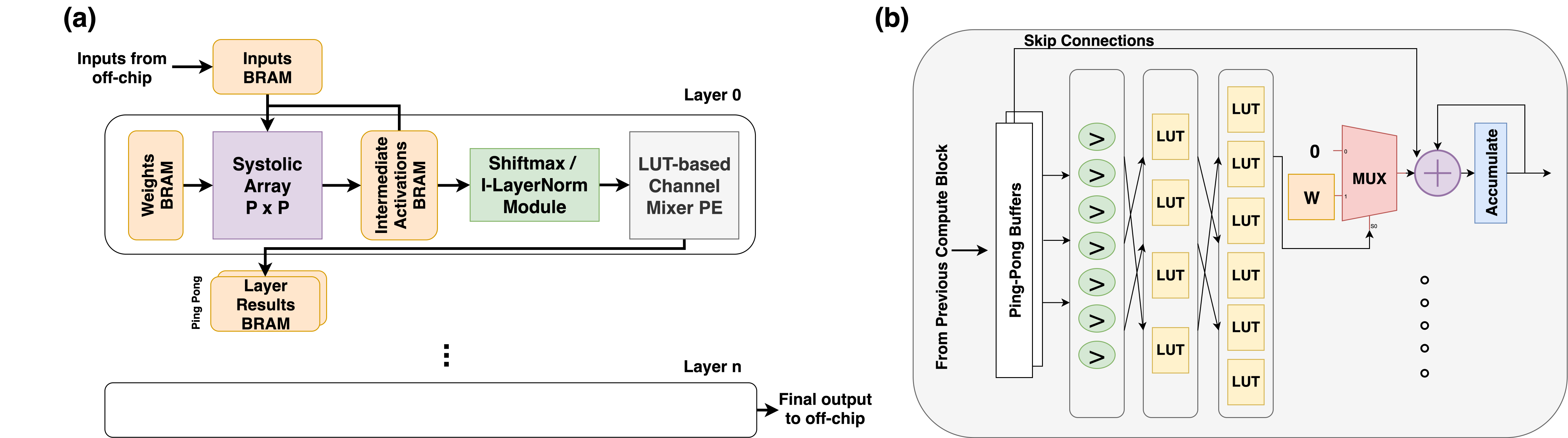}}
  \caption{Hardware Accelerator Design for \ours -- (a) the overall design (b) the PE design for the LUT-based channel mixer.}
  \vspace{-1mm}
  \label{fig:hw_design}
\end{figure*}

\subsection{FPGA Acceleration}
\label{sec:pe}

\noindent\textbf{Channel Mixer PE Design: }For the \ours accelerator design, we propose a dedicated processing element (PE) block design for the LUT-based channel mixer, shown in Fig.~\ref{fig:hw_design}(b). 
As the connections between two layers could be any arbitrarily learnt mapping, all output bits from the previous layer would be required to form input addresses for the next layer. Hence, to avoid stalls, we design this PE to be parallely processing all elements along the channel-dimension for a given row of input activations, with subsequent rows processed through the PE in a pipelined fashion. 
Ping-pong buffers at the input of the PE are used to stage the row of input activations currently being processed in one buffer, while the other buffer is simultaneously filled with the next row. 
The thermometer encodings are realized using comparator blocks, and the LUT neurons in the feed-forward LUT layers 
are implemented using the LUTs within the logic slices of the FPGA. 
For the conditional summation layer, we introduce an adder circuit preceded by a 2:1 MUX for each dimension in the latent space. For the specific channel, the MUX and adder circuit iteratively adds the contributions from all the LUTs in the previous layer. Note that while we could also introduce an adder tree for each of these for low-latency, this would increase the resource consumption quite significantly. 
Furthermore, we find that the overall throughput of the network would be bottlenecked by other slower layers, and hence this iterative approach allows us to time-balance the channel mixer stage with the other stages of the network, while conserving resources. 
This adder circuit also accumulates the skip connections from the ping-pong buffer. 

\noindent\textbf{Overall Design: }
Fig.~\ref{fig:hw_design}(a) shows the overall \ours accelerator design. To offer high throughput, we design dedicated blocks for implementing each layer in the encoder stack, and multiple frames are pipelined across these layers. The inputs are written into the first layer, and outputs are read back from the last layer. All model weights are read and stored in dedicated BRAMs at initialization, and remain stationary on-chip. The matrix multiplications in the MHA token-mixer are catered to by a $P \times P$ systolic array, as used in literature typically \cite{mlpsystolic}. This is interfaced with the PE block designed for the channel mixer. Dedicated blocks are implemented for computing the integer-based non-linear operations, namely, I-LayerNorm and ShiftMax, as proposed in I-ViT~\cite{ivit}.

As mentioned in Sec.~\ref{sec:intro}, an important aspect to note here is that by introducing LUT-based layers instead of conventional MLP layers, we are not trading off compute with additional memory. On FPGAs, 
compute blocks are majorly built up of DSPs and LUTs. Consequently, any processing element for conventional neural network acceleration gets mapped to these logic elements on device. In our proposed model, the LUTs in the channel mixer would get directly mapped to these LUTs on device, incurring no additional overhead on the memory. Infact on the contrary, the LUT-based implementation eliminates the memory requirement associated with those weights that would have been present in a conventional layer; while also eliminating the associated off-chip memory access. 

\section{Evaluation} \label{sec:eval}
\subsection{Experimental Setup} \label{sec:exp}


\noindent\textbf{Model Backbone \& Datasets:} As the goal of our work is to develop an edge-optimized deployment solution for vision transformers, we consider I-ViT-T~\cite{ivit}, an \textbf{INT8 quantized} model as our \textbf{baseline}. To evaluate the performance of our proposed channel mixer and model architecture changes, we consider three datasets for evaluation -- CIFAR-10 \cite{cifar10}, CIFAR-100, Flowers-102~\cite{flowers} and TinyImageNet \cite{tinyimagenet}. We particularly choose these medium-sized datasets considering that we are optimizing these models for resource-constrained environments, and inline with prior works on LUT-neuron based networks and edge efficient models~\cite{dwn, finn}. I-ViT-T is a fully-quantized INT8 precision model, and designing \ours with this backbone allows us to gauge our 
design's benefits in optimizing an already quantized model. For consistency with other ViTs, we resize the images in these datasets to $224 \times 224$ and apply the same data augmentations used in DeiT~\cite{deit}. 

\noindent\textbf{Channel Mixer Configuration:} As described in Section \ref{sec:channelmixer}, the channel mixer allows the following configurations to be specified -- 
the number of bits of thermometer encoding, the number of layers of LUTs, the number of LUTs in each layer, and the {output precision for the encoded values}. The number of inputs, and the output channel dimensions would be determined by the model backbone to which the channel mixer is being integrated. The configurable parameters are indicative of the complexity of the channel mixer, and these can be scaled based on the baseline vision transformer variant it is being integrated to. With our experimental observations, we found a two LUT-layer channel mixer configuration, with $(768,192)$ LUTs in the layers respectively, with a $8$-bit thermometer encoding to be optimal for the I-ViT-T model backbone. Further, \textbf{post-training}, the \textbf{encoded values} in the conditional summation layer were \textbf{quantized to 4-bit}. The parameters were chosen using the configurations of the original MLP in the baseline as a guidance, with effectively the same number of LUT neurons being used in each layer as in the MLP. 

\noindent\textbf{Training Methodology:} We define the LUT-based channel mixers using custom PyTorch classes, and integrate this with the MHA based token mixer and train the resultant \ours. We use the default train-test split that these datasets come with, train these models on a A100 GPU, and primarily use the same learning rate, scheduler and optimizer as the baseline. 

\begin{table*}[htbp!]
\centering
\vspace{-3mm}

\begin{tabular}{ p{2.6cm}p{1.4cm}p{1.4cm}p{1.75cm}p{1.7cm}p{1.75cm}p{2cm}p{2cm}}
 \toprule
 Work & CIFAR-10 & CIFAR-100 & Tiny-ImageNet & Flowers-102 & Model Size & Inference Latency & Energy/ Inference \\
 \midrule
 I-ViT-T (baseline, int8)  & 95.4\% & 79.2\% &  60.4\% & {91.3\%}
 & 5.06 MB & 6.93 ms & {4.05 mJ} \\
 \textbf{\ours (ours, int8)} & 95.5\% & 78.8\% & 60.9\% & {91.6\%}
 & 1.93 MB & 5.33 ms & {2.14 mJ} \\
 \bottomrule
\end{tabular}
\vspace{5pt}
\caption {Inference performance comparison of \ours ~against I-ViT-T, a fully quantized ViT baseline on the target FPGA. On CIFAR-10, CIFAR-100, Tiny-ImageNet and Flowers-102 datasets, \ours achieves similar accuracies as the baseline. 
}

\label{table:acc}
\end{table*}

\noindent\textbf{FPGA Evaluation:} 
To evaluate the inference efficiency, we generate SystemVerilog RTL \cite{system-verilog} for the accelerator design, 
with the RTL code for the PE generated from the trained model using custom mako scripts. For our evaluation, we consider the systolic array to be of dimensions $P \times P = 32 \times 32$. {For the baseline I-ViT-T model acceleration, we consider a pipelined $32 \times 32$ systolic array for the MLP and MHA blocks.} 
We synthesize our design on AMD Xilinx xcvu9p-flgb2104-2-i (Virtex series) FPGA, with a target clock frequency of 200 MHz using Vivado Design Suite. For power and energy estimation, we use a default switching activity factor of 12.5\% and consider the total device power using AMD Power Design Manager (PDM)~\cite{amd_pdm}. 
We evaluate the overall model's performance in terms of latency, and energy consumed per image inference. 

\vspace{-1mm}

\subsection{Results}

\noindent\textbf{Model Accuracy:} 
We trained our model, and performed a post-training quantization on the encoded value in the conditional summation layers to int4 precision. As noted in Table~\ref{table:acc}, with CIFAR-10, \ours achieves an accuracy of 95.5\% against the baseline (I-ViT-T) model accuracy of 95.4\%, 78.8\% accuracy on CIFAR-100, 60.9\% on Tiny-ImageNet, and 91.3\% on Flowers-102 dataset, with a 60\% smaller model size. 
These findings suggest that \ours performs comparably to the baseline in terms of model accuracy. Table~\ref{table:acc-only} compares our model's performance in terms of accuracy, parameter count, and number of MAC operations against other common iso-accuracy vision transformer models. \ours achieves better accuracy than most other works, while involving fewer parameters and less than 50\% MAC operations compared to the baseline. We also note that \ours performs much better compared to prior works like FINN, and LUT-neuron based works, that report accuracy on CIFAR-10 in the range of 40-80\% (Table~\ref{table:wnn-acc-only}). 


\begin{table}[htbp!]
\centering

\begin{tabular}{ p{2.6cm}p{1.8cm}p{1.2cm}p{1.25cm}}
 \toprule
 Work & CIFAR-10 Top~1~Accuracy & \#Params (M) & \#GMACs/ inference \\
 \midrule
 CCT-2/3x2 \cite{cct}  & 89.7\% & 0.28 & 0.04  \\
 CCT-7/3x2 \cite{cct}  & 95.0\% & 3.85 & 0.29  \\
 CCT-7/3x1 \cite{cct}  & 96.5\% & 3.76 & 1.19  \\
 DeiT-T \cite{deit}  & 94.8\% & 5.3 & 1.31  \\
 I-ViT-T~\cite{ivit} (baseline)  & 95.4\% & 5.3 & 1.31  \\
 \textbf{\ours (ours)}  & 95.5\% & 2.5 &  0.65 \\
 
 \bottomrule
\end{tabular}
\caption {Comparison of optimized ViTs against \ours. 
}

\label{table:acc-only}
\end{table}

\begin{table}[htbp!]
\centering
\vspace{-3mm}

\begin{tabular}{ p{2cm}p{1.8cm}p{1.2cm}p{0.8cm}p{0.9cm}}
 \toprule
 Work & CIFAR-10 Top~1~Accuracy & Param Size (KiB) & LUTs (K) & Energy/ inf (mJ) \\
 \midrule
 TreeLUT~\cite{treelut} & 42.9\% & -- & 2.2 & -- \\
 DiffLogicNet~\cite{petersen2022}  & 57.3\% & 250 & 283.3 & -- \\
 DWN \cite{dwn}  & 57.5\% & 23.4 &  45.7 & 0.004 \\
 FINN \cite{finn}  & 80.1\% & 183.1 & 46.3 & 0.15 \\
 \textbf{\ours (ours)}  & 95.5\% & 1976.3 & 587 & {2.14} \\
 
 \bottomrule
\end{tabular}
\caption {Comparison of \ours against prior works on LUT-based models -- none of these  implement a ViT architecture, and the accuracies are considerably low. This demonstrates how our work advances the performance of LUT-based models on vision tasks with a ViT model backbone.  
}

\label{table:wnn-acc-only}
\end{table}

\noindent\textbf{Optimizing the Optimized:} We also demonstrate how our proposed design further optimizes the tiniest variant of Compact Convolutional Transformers \cite{cct} -- a model that is already optimized for small size, and well-suited for 
resource-constrained environments. With a \ours design built on the CCT-2/3x2 backbone, we achieve an accuracy of 87.4\% on the CIFAR-10 dataset while still reducing MAC operations by about $1.5 \times$. 

\noindent\textbf{Energy Efficiency: } We evaluate the energy consumption for inference of a single image for our design on the target device. 
As shown in Table~\ref{table:acc}, \ours achieves a $1.9 \times$ improvement in energy efficiency (energy per inference) over a fully-quantized baseline accelerator implemented on the same FPGA. {Note that this is a conservative estimate, as it does not account for the additional energy spent in the baseline implementation to fetch MLP weights that cannot be accommodated on-chip.}

\noindent\textbf{Latency: } As noted in Section \ref{sec:pe}, our Processing Element is designed such that it avoids stalls due to the varied computations in the token-mixer and channel-mixer blocks. As a consequence, our design is not fully optimized for latency, and we primarily target energy efficiency. Our analysis shows that we still achieve a $1.3 \times$ improvement in latency by virtue of our optimized channel mixer, as reported in Table~\ref{table:acc}.

\noindent\textbf{Comparison with prior ViT accelerators: } Table~\ref{table:comp-prior} compares the FPGA acceleration of \ours against other prior works on ViT accelerators. For comparable resource usage, we use systolic arrays of 16$\times$16 dimensions. As seen, with a low power edge, our approach achieves superior efficiency (FPS/W) and a high FPS compared to most prior works. HG-PIPE~\cite{hgpipe}, the only exception, consumes 4$\times$ the power as \ours. We note that some of these works, including HG-PIPE, are aggressively quantized (3-bit) -- we demonstrate superior performance even with conservative 8-bit quantization scheme for our design. With our optimizations, \ours fits within the on-chip memory.

\noindent\textbf{Comparison with other Multiplication-free works: } 
In Fig. \ref{fig:mult-free},
we evaluate our LUT-based channel mixer's efficacy against some popular multiplication-free and approximate matrix-multiplication works that have been gathering traction of late; including Affine \cite{affine} and LUT-NN \cite{lut-nn}, in terms of the computations involved. For a channel mixer with the same number of neurons, while these works eliminate multiplications over the baseline, they do it at the cost of increased reads or adds, as each multiplication is replaced by an equivalent add or look-up operation. On the contrary, with learned LUTs, \ours achieves similar model performance while reducing the number of lookup and add operations by a significant factor -- demonstrating the efficacy of learning LUTs while training. 

\begin{figure}[htbp!]
  \centering
\vspace{-5pt}  {\includegraphics[width=0.35\textwidth, keepaspectratio]{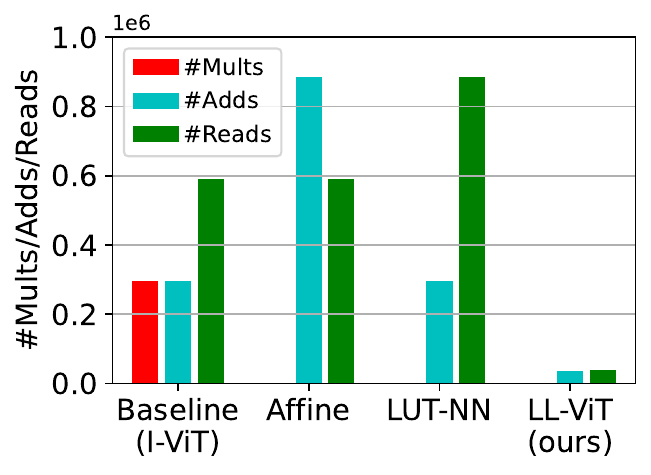}}
  \caption{Comparison of \ours against multiplication-free designs in terms of the number of multiplications, additions, and read operations in the channel mixer block.}
  \vspace{-3mm}
  \label{fig:mult-free}
\end{figure}

\begin{table*}[htbp!]
\centering
\vspace{-3mm}

\begin{tabular}{ p{3.2cm}p{1.5cm}p{1.3cm}p{1.3cm}p{1.3cm}p{1.3cm}p{1.3cm}p{1.3cm}p{1.3cm}}
 \toprule
 Work & Platform & DSPs & LUTs (K) & FFs (K) & BRAM36s & FPS & Power(W) & Efficiency (FPS/W) \\ 
 \midrule
 ViTA (ISCAS 23) \cite{vita} & Pynq Z1 & 0 & 22.8 & 5 & * & 19 & 0.88 & 21.60 \\
 HG-PIPE (ICCAD 24) \cite{hgpipe} & VCK190 & 312 & 669 & * & 1006 & 7118 & 46.7 & 152.42 \\
 HeaT-ViT (HPCA 23) \cite{heatvit} & ZCU102 & 1968 & 137 & 126 & 355 & 183 & 9.45 & 19.40 \\
 Huang et al. (TCAS-I) \cite{huang-tcas} & ZCU102 & 1268 & 114 & 168 & 648 & 245 & 29.6 & 8.27 \\
 SSR (FPGA 24) \cite{ssr} & VCK190 & 14405* & 619 & 849 & 1456 & 4545 & 46 & 98.8 \\
 ME-ViT (HiPC 24)~\cite{mevit} & Alveo200 & 1024 & 192 & 132 & 288 & 352 & 9.3 & 37.87 \\
 \textbf{LL-ViT (ours)} & Virtex & 0 & 589 & 229 & 1425 & {1083} & {10.9} & {99.35} \\
 \bottomrule
\end{tabular}
\vspace{5pt}
\caption {Comparison of our work against other prior works for IViT-T or equivalent models. (* = approximated/ not reported) 
}

\label{table:comp-prior}
\end{table*}

\begin{table}[t!]

\centering
\newcolumntype{C}[1]{{\centering}m{#1}}
\vskip 0.1in

\vspace{-9pt}
\centering
\vskip 0.1in
\begin{tabular}{ ccccc}
 \toprule
 Stage & Baseline I-ViT-T & \ours \\ 
 \midrule
 $Q$, $K$, $V$ & 23.59 uJ (each) & 23.59 uJ (each) \\ 
 $QK^T$ & 27.52 uJ & 27.52 uJ \\ 
 $S.V$ & 27.52 uJ & 27.52 uJ \\ 
 Multi-Head Concat & 23.59 uJ & 23.59 uJ \\ 
 MLP Dense 1 & 94.35 uJ & \\ 
 MLP Dense 2 & 94.35 uJ & \multirow{-2}{*}{28.8 uJ} \\ 
 \midrule
 \textbf{Total} & \textbf{338.10 uJ} & \textbf{178.2 uJ} \\ 
 \bottomrule
\end{tabular}
\vspace{5pt}
\caption{Comparing the layerwise breakdown of energy consumption between the baseline and \ours ~for a single encoder layer, per sample inference. A 32x32 systolic array is used for the baseline model and for the non-weightless layers in \ours. We ignore the energy consumption of the other components including SoftMax, GELU and LayerNorm for clarity in analysis, as these were found to be minimal. 
} \label{table:end-to-end}

\end{table}


\subsection{Ablation Studies}\label{sec:hwperf}
\noindent\textbf{Quantized Baselines: } We consider various quantization schemes for the baseline vision transformer. As shown in Fig. \ref{fig:energy-precision}, \ours designed with the quantized backbone consistently demonstrates an energy efficiency of over $1.8\times$ regardless of the quantization scheme used \textbf{(4-bit/ 2-bit/ binary)}.

\noindent\textbf{Scaling Trends: } We also show that energy efficiency offered by \ours scales comparably with the latent dimensions of the network (Fig. \ref{fig:ablation}(a)) and the image size (Fig. \ref{fig:ablation}(b)). This suggests that as we continue to scale LL-ViTs to larger or smaller models, they would consistently offer the energy savings reported.  

\noindent\textbf{Layerwise Energy Breakdown: } Table \ref{table:end-to-end} indicates the breakdown of energy consumption in a single encoder block of \ours against a fully-quantized baseline I-ViT design. This clearly indicates that \ours eliminates the most energy intensive blocks of the vision transformer (MLP), and introduces an alternate energy-efficient channel mixer block. 

\noindent\textbf{Encoded Value Post-training Quantization: } As mentioned in Sec. \ref{sec:channelmixer}, the encoded value in the channel mixer can be quantized post-training to match the precision desired by the rest of the network. We observe the overall accuracy of \ours to be 95.6\% at int8 quantization, 95.5\% at int4 quantization, and 90.3\% at int2 quantization. As int4 quantization offers optimal tradeoff, we stick to it for all our evaluations.

\begin{figure}[htbp!]
  \centering
  {\includegraphics[width=0.35\textwidth, keepaspectratio]{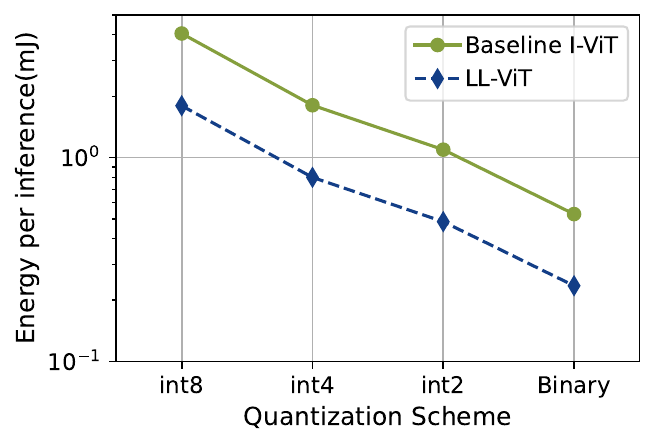}}
  \caption{Energy per inference of baseline I-ViT vs. \ours for varying quantization schemes.}
  \vspace{-1mm}
  \label{fig:energy-precision}
\end{figure}

\begin{figure}[htp]
    \centering
    \subfloat[\centering]{\includegraphics[width=4.25cm]{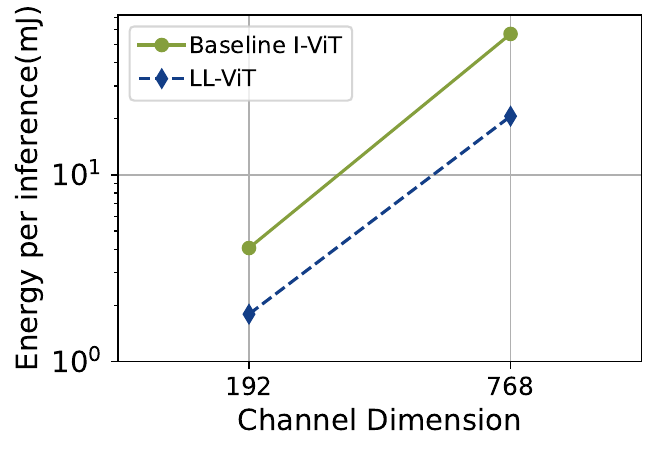} }%
    \subfloat[\centering]{\includegraphics[width=4.25cm]{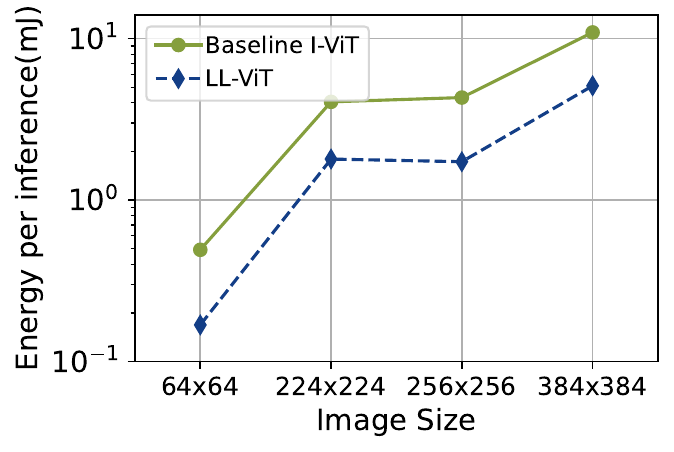} }%
    \vspace{-0.075in}
    \caption{(a) Energy per inference of the baseline {I-ViT} vs. \ours with varying latent dimensions (with the same number of neurons in both models). (b) Energy per inference of the baseline I-ViT vs. \ours with varying image sizes, with the model configuration remaining the same.}%
    \label{fig:ablation}%
\end{figure}

\section{Discussion} \label{sec:discuss}

While we primarily demonstrate our work with a tiny ViT baseline (I-ViT-T) as we specifically target edge applications of small tasks and datasets, the technique proposed with \ours could also be applied to larger variants of vision transformers. 
The hardware performance improvements over the baseline, would be similar for any vision transformer model, considering the structural composition of the accelerator, as shown in the case of CCT. 
Similarly, we also note that while our primary baseline was INT8 quantized, the proposed technique can be integrated with complementary works on aggressive quantization  like BinaryViT~\cite{binaryvit}, and still achieve performance improvement, as alluded in Fig.~\ref{fig:energy-precision}. We view this work as a stepping stone towards a class of learnable LUT-based tiny transformer models, 
that are competitive to the current energy-inefficient transformers. 

\vspace{-1mm}
\section{Conclusion} \label{sec:conclusion}
In this work we introduce Learned-LUT based Vision Transformers (\ours), an effort 
to develop edge-efficient vision transformers targeted for FPGA acceleration. 
We identify an opportunity to reduce computational and memory demands by targeting channel mixers, replace these with the proposed LUT-based channel mixers, and design an accelerator for the model. 
Based on our model and performance evaluations, we report
 $1.9 \times$ energy efficiency, and $1.3 \times$ lower latency against quantized baselines at comparable model accuracies. We also reduce the model size by 60\%, enabling the remaining model weights to be fully stationary on-chip. These results illustrate that LL-ViTs are well-positioned as a promising lightweight alternative to traditional vision transformers -- paving the way for tiny ViTs deployable at the edge. Although LUT-neuron based models have been applied to many small-scale problems in the past, to the best of our knowledge, this is the first time that its usefulness in constructing a vision transformer model has been demonstrated. 

\section*{Acknowledgements}
This research was supported by Semiconductor Research Corporation (SRC) Task 3148.001, National Science Foundation (NSF) Grants \#2326894, \#2425655 (supported in part by the federal agency and Intel, Micron, Samsung, and Ericsson through the FuSe2 program), NVIDIA Applied Research Accelerator Program, and compute resources on the Vista GPU cluster through CGAI \& TACC at UT Austin. Any opinions, findings, conclusions, or recommendations are those of the
authors and not of the funding agencies.

\bibliographystyle{IEEEtran}
\bibliography{mlsys_paper}



\end{document}